\documentclass[10pt,twocolumn,letterpaper]{article}

\usepackage{iccv}
\usepackage{times}
\usepackage{epsfig}
\usepackage{graphicx}
\usepackage{amsmath}
\usepackage{amssymb}
\usepackage{algorithm}
\usepackage{algorithmicx}
\usepackage{algpseudocode}
\usepackage{subfigure}
\usepackage{multicol}
\usepackage[font=it]{caption}
\usepackage{paralist}
\usepackage{setspace}
\usepackage[pagebackref=true,breaklinks=true,letterpaper=true,colorlinks,bookmarks=false]{hyperref}
\usepackage{indentfirst}

\usepackage[final]{pdfpages}
\setboolean{@twoside}{false}

\makeatletter
\let\ps@plain\ps@empty
\let\ps@headings\ps@empty
\makeatother

\usepackage[breaklinks=true,bookmarks=false]{hyperref}

\DeclareMathOperator*{\argmax}{argmax}

\usepackage[breaklinks=true,bookmarks=false]{hyperref}

\iccvfinalcopy 

\def\iccvPaperID{2931} 

\ificcvfinal\fi
\begin{document}

\title{3DCNN-DQN-RNN: A Deep Reinforcement Learning Framework for Semantic Parsing of Large-scale 3D Point Clouds}

\author{
	Fangyu Liu\thanks{Equal contribution.}  {\small $\ ^1$} , 
	Shuaipeng Li\footnotemark[1]  {\small $ \ ^1$} , 
Liqiang Zhang\thanks{Corresponding authors are Liqiang Zhang and Chenghu Zhou, E-mail:\{\tt zhanglq@bnu.edu.cn; zhouch@lreis.ac.cn\}.}  {\small $\ ^1$}, 
Chenghu Zhou\footnotemark[2] {\small $\ ^2$} , \\
Rongtian Ye{\small \ $^1$} , Yuebin Wang{\small $\ ^1$} , 
Jiwen Lu{\small $\ ^3$} \\
{\small $^1$Faculty of Geographical Science, Beijing Normal University, Beijing, China} \\
{\small  $^2$Institute of Geographic Sciences and Natural Resources Research, Chinese Academy of Sciences} \\
{\small  $^3$Department of Automation, Tsinghua University, Beijing, China} \\
}

\maketitle

\begin{abstract}
Semantic parsing of large-scale 3D point clouds is an important research topic in computer vision and remote sensing fields. Most existing approaches utilize hand-crafted features for each modality independently and combine them in a heuristic manner. They often fail to consider the consistency and complementary information among features adequately, which makes them difficult to capture high-level semantic structures. The features learned by most of the current deep learning methods can obtain high-quality image classification results. However, these methods are hard to be applied to recognize 3D point clouds due to unorganized distribution and various point density of data. In this paper, we propose a 3DCNN-DQN-RNN method which fuses the 3D convolutional neural network (CNN), Deep Q-Network (DQN) and Residual recurrent neural network (RNN) for an efficient semantic parsing of large-scale 3D point clouds.  In our method, an eye window under control of the 3D CNN and DQN can localize and segment the points of the object's class efficiently. The 3D CNN and Residual RNN further extract robust and discriminative features of the points in the eye window, and thus greatly enhance the parsing accuracy of large-scale point clouds. Our method provides an automatic process that maps the raw data to the classification results. It also integrates object localization, segmentation and classification into one framework. Experimental results demonstrate that the proposed method outperforms the state-of-the-art point cloud classification methods.
\end{abstract}
\section{Introduction}
In recent years, deep learning techniques have had a great success in image, speech and text recognition. However, few studies have focused on 3D large-scale point cloud classification. Different from images whose spatial relationships among pixels can be caught by sliding windows, the points in a point cloud are unorganized and the point density is uneven. It is a challenge to accurately parse an unorganized and unoriented 3D point cloud corrupted with noise, outliers, and under-sampling. 

\begin{figure}[t]
\begin{center}
   \includegraphics[width=0.9\linewidth]{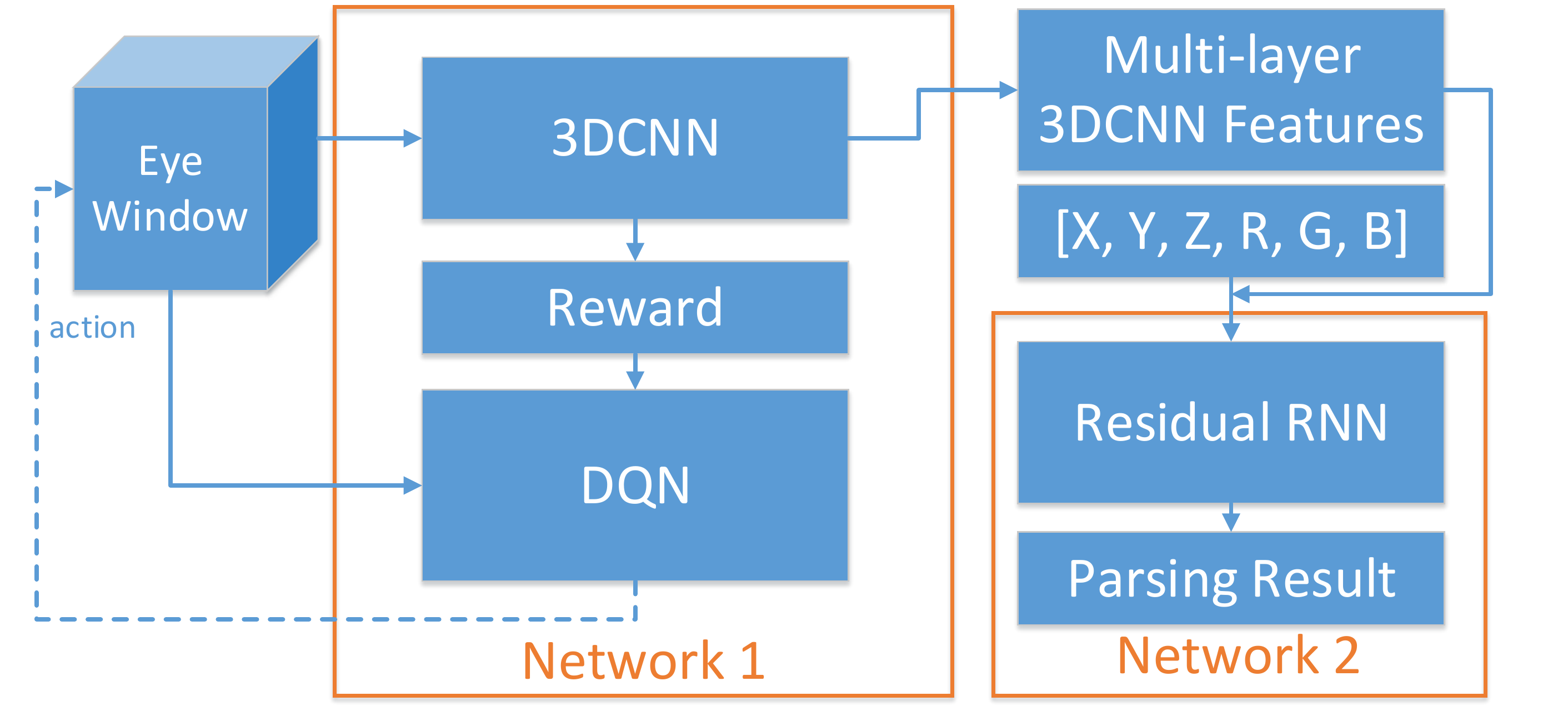}
\end{center}
   \caption{The framework of the proposed approach. $(X,Y,Z)$ and $(R, G, B)$ represent
   3D coordinate and RGB colors of each point $P_i$ in the original point cloud.}
\end{figure}
In this paper, we propose a deep reinforcement learning method, \ie 3DCNN-DQN-RNN, to automatically parse large-scale 3D point clouds. Through supervised learning, the 3D convolutional neural network (CNN) has the ability to learn features about shape, spatial relationship, color and context of the points in the point cloud from multiple scales, and then encode them into a discriminative feature representation called 3D CNN feature. To recognize a certain object from the point cloud, an eye window traverses the whole data for localizing the points belonging to a certain class. The size and position of the eye window are controlled by the Deep Q-Network (DQN). During the traversal, the DQN gets the probability (computed through the reward vector output by the 3D CNN) that the eye window contains the class object. It next determines which region is worth looking at, and then makes the eye window change its size and move towards the region. We re-apply the 3D CNN to compute the reward vector of the points in the new eye window. Repeat the process until the eye window accurately envelops the points of the class object. Once the class object is localized and enveloped, the 3D CNN feature, coordinate and color of each point within the eye window are combined into a vector which is input into the Residual recurrent neural network (RNN). The Residual RNN further abstracts and merges feature representations of the points in the eye window. The output of the residual RNN is the classification result. The eye window keeps moving until all the points of the class objects are parsed.
The main contributions of our work are as follows:
\begin{enumerate}[(i)]
\item We propose a novel deep reinforcement learning model to precisely parse large-scale 3D point clouds. Most of the parameters in the 3DCNN-DQN-RNN are learned, and thus the intensive parameter tuning cost is significantly reduced. Moreover, it integrates object localization, detection, retrieval and classification into one framework.
\item The Residual RNN abstracts and merges 3DCNN feature, coordinate and color of each point from multi-scales into a more robust and discriminative feature representation. High-quality classification performance is achieved.
\item The eye window under the control of the DQN is applied to localize and segment class objects. The localization and segmentation are accurate, automatic and less expensive.

\end{enumerate}

\section{Related Work}
For parsing point clouds, many recent methods used classifiers trained on hand-crafted features like Spin Images~\cite{johnson1999using,zhang2016multilevel}, eigenvalues~\cite{wang2015multiscale}, contextual features~\cite{yang2017computing} or specific color, shape and geometry features~\cite{martinovic20153d}. Usually, spectral-spatial features~\cite{zhang2013tensor} or spectral features~\cite{zhang2006pixel} are difficult to be obtained due to lack of spectral information in point clouds. For example, Chehata \emph{et al.} ~\cite{chehata2009airborne} used Random Forests trained on 21 features to classify 3D point clouds into 5 classes. Kragh \emph{et al.} utilized the SVM classifier with 13 features to classify point clouds~\cite{kragh2015object}. Lafarge and Mallet distinguished four classes of interest, \ie building, vegetation, ground, and clutter, from 3D point clouds of urban environments based on four different features~\cite{lafarge2012creating}. Zhou and Neumann classified points in the urban point cloud into trees, buildings, and ground through the 2.5D characteristic criterion~\cite{zhou2013complete}. Wang \emph{et al.} clustered the point cloud into multi-levels, and derived the point cluster features from the point-based feature descriptors. The Adboost Classifier is applied to classify the point cluster with the finest level into semantic classes~\cite{wang2015multiscale}. These approaches fail to adequately utilize the consistency and complementary information between features, which are difficult-to-capture high-level semantic structures.

Recently, deep learning techniques have been applied to 3D object recognition tasks on 3D data like RGBD images and point clouds. The techniques can automatically learn features from 3D data. Wu \emph{et al.} presented a volumetric CNN architecture on 3D voxel grids to represent a geometric 3D shape for object classification and retrieval~\cite{wu20153d}. Zhu \emph{et al.} used depth images with different perspectives of 3D objects as the input, and then utilized auto-encoder with pre-training using DBN to extract features~\cite{koppula2011semantic}. Only a few studies have applied deep learning techniques in point cloud classification. Guan \emph{et al.} classified 10 species of trees by using the DBN for the vertical profile of the tree point clouds~\cite{guan2015deep}. Based on a 2D convolutional neural network, Maturana and Scherer proposed a 3D CNN for object binary classification task based on LiDAR data~\cite{maturana20153d}. Later, they further introduced 3D CNNs for landing zone detection from LiDAR data~\cite{maturana2015voxnet}. To tackle a more general object recognition task with LiDAR and RGBD point clouds from different modalities, different representations of occupancy were proposed in ~\cite{maturana2015voxnet}. A volumetric occupancy grid representation and a supervised 3D CNN are integrated to improve the performance. To make 3D CNN architectures fully exploit the power of 3D representations, Qi \emph{et al.} introduced two distinct network architectures of volumetric CNNs for object classification on 3D data~\cite{qi2016volumetric} . These methods work well on object detection problems but they are hard to parse large scale point clouds directly. There are also detection-based semantic parsing methods for point clouds or RGBD data. Song \emph{et al.} presented a 3D ConvNet pipeline for amodal 3D object detection including a region proposal network and a joint 2D+3D object recognition network~\cite{song2016deep}. Armeni \emph{et al.} proposed a sliding window approach which combines the local and global features of the object for semantic parsing of the indoor point cloud~\cite{armeni20163d}. Both approaches require heavy hand-crafted work like designing suitable sliding windows or searching boxes. In~\cite{qi2016pointnet}, a deep neural network called PointNet was introduced to perform 3D shape classification, shape part segmentation and scene semantic parsing tasks on point cloud.

For object segmentation, most works deal with 2D/2.5D image-based segmentation. Long \emph{et al.} introduced a fully connected network (FCN)-based end-to-end framework in which the trained CNN can perform pixel-wise prediction of class of images~\cite{long2015fully}. Chen \emph{et al.} presented the DeepLab ~\cite{chen2014semantic} which were later developed with the help of Atrous Spatial Pyramid Pooling and a combination with CRFs. A similar structure called dilated convolution was introduced by Yu and Koltun to extend the receptive fields~\cite{yu2015multi}.  Studies attempting to apply the FCN in high-dimensional data have shown in recent publications. For example, Song \emph{et al.} introduced SSCNet~\cite{song2016ssc} which developed the work of~\cite{song2014sliding} to semantically segment RGBD images. However, these methods are hard to be applied to segment large-scale 3D point clouds with significant data missing.

\section{Proposed Approach}
Humans usually solve vision problems through a harmonious combination of reinforcement learning and hierarchical sensory processing systems~\cite{fukushima1980neocognitron,serre2005object}. Experts expect the progress in computer vision in future to come from systems that are trained end-to-end and combine ConvNets with RNNs using reinforcement learning to decide where to look~\cite{lecun2015deep}. In this paper, we apply this idea to parse large-scale point clouds. Like the vision behavior patterns, they usually recognize an object from a complex scene as follows. As a first step, they roughly look at the whole scene and find the approximate location of the target. Next, they focus on the object and separate it from the background. Imitating this behavior, we propose a deep reinforcement learning framework for semantically parsing large-scale 3D point clouds through recognizing every class object which is illustrated in Figure 1.

\begin{figure}
\begin{center}
   \includegraphics[width=1.0\linewidth]{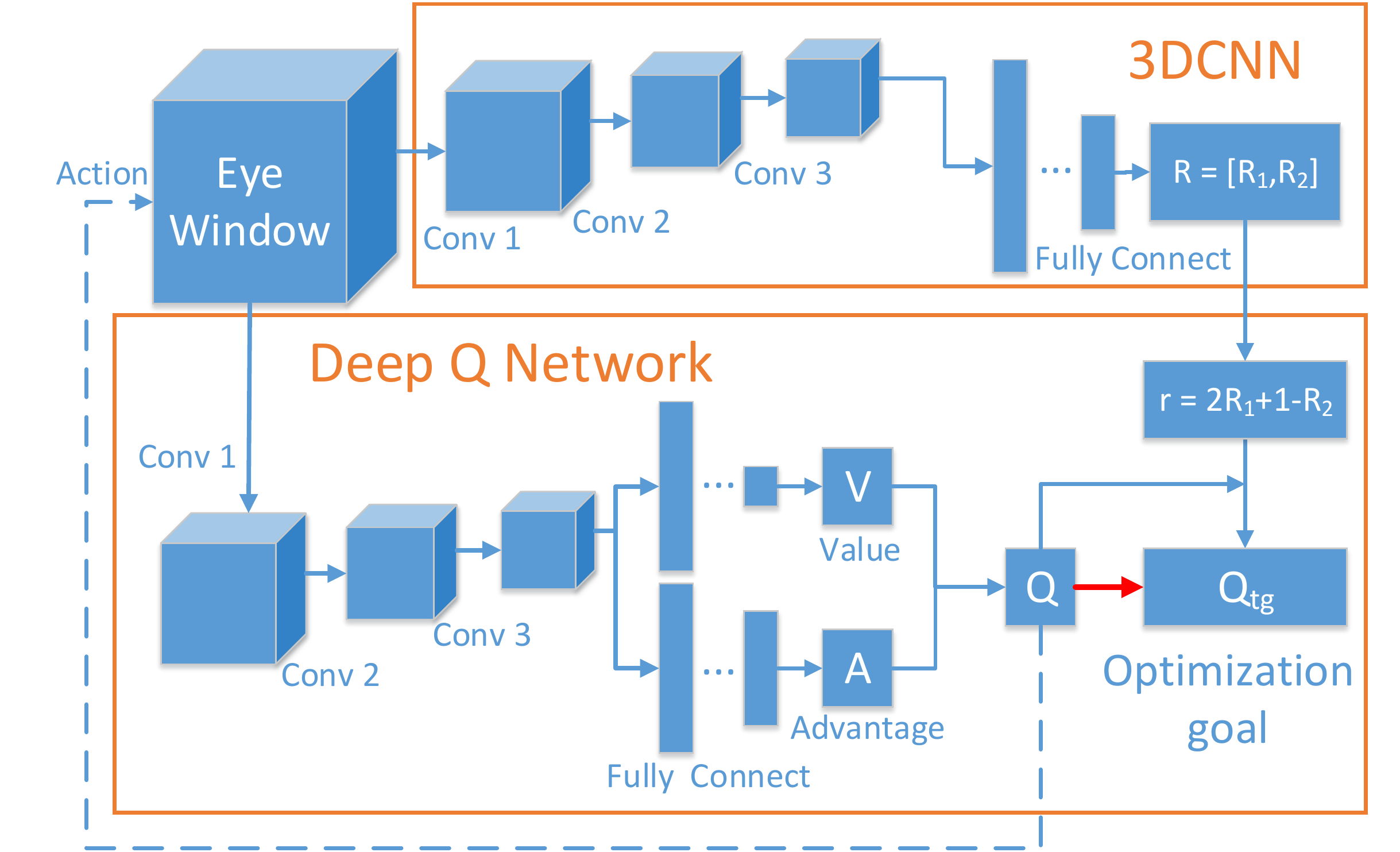}
\end{center}
  \caption{Structure of Network 1.}
\end{figure}
\subsection{Network 1: Recognition and Localization through 3DCNN-DQN}
The first part of Network 1 is a 3D CNN. It is trained to determine different class objects. In the meantime, the feature representation of the points of each class are obtained by the 3D CNN. These features are then encoded and are applied in the following networks. The second part of Network 1 is a DQN whose goal is to detect and localize the objects. Based on the feedback from the 3D CNN, the DQN automatically perceives the scene and adjusts its parameters to localize the objects.

To parse the point cloud efficiently and effectively, we first set an eye window with a certain size and make it move in the scene, and then apply the 3D CNN to recognize the data inside the eye window. The DQN gets the probability (computed through the reward vector output by the 3D CNN) that the eye window contains the target class object. It next determines which region is worth looking at, and then makes the eye window change its size and move towards the region. We re-apply the 3D CNN to get the reward vector of the points in the new eye window. Repeat the process until the eye window accurately envelops the points of the class objects. Finally, the 3D CNN features and all the points within the eye window are taken as the input of Network 2. Figure 2 shows the 3DCNN-DQN structure.

\subsubsection{Learning Multi-Scale Features Using the 3D CNN}
Objects in large-scale scenes are various in size, shape and position. However, the structure information and spatial relationship of different objects can be represented from different scales. To achieve a more purified classification result, we do not directly make use of the 3D CNN to parse the point clouds. Instead, we compute the probability of a certain class being contained inside the eye window after the multi-scale convolution is performed. The probability is indicated in the output reward vector $[R_1,R_2]$ and calculated using the following formula:
\begin{equation}
f(E(p,q))=\theta^\top E(p,q)=2R_1+1-R_2=r
\end{equation}
where $f$ is a confidence function; $R_1$ is the first dimension of the reward vector which is the output of the 3D CNN; $R_2$ is the second dimension of the reward vector; $\theta$ is the parameter of the 3D CNN; $E$ represents all the points within the eye window whose size is $q$ and position is $p$. Both the reward vector and the points within the eye window are taken as the input of the DQN. Every time the DQN receives the input from the 3D CNN, it updates the size and position of the eye window. This procedure does not stop until the eye window localizes the target class object precisely, and then every convolutional layer outputs a feature. All the features are taken as the 3D CNN feature vector of all the points within the eye window.

\paragraph{a. Converting the Point Cloud to Voxel Grid } 
3D Point clouds are usually unorganized, which makes the CNNs difficult to extract features from them. We first convert the point cloud to 3D voxel grids (occupancy grids), and then segment the grids into o dense, regular and ordered small units. Each unit is given a sole attribute $(X; Y; Z; R; G; B)$. The coordinate $X,Y,Z$ of each unit represents the position of the unit in the voxel grids, and $R,G,B$ represent the average color value of the points in the unit.

\paragraph{b. 3D CNN Structure} 
The 3D CNN contains three convolutional layers and multiple fully-connected layers. The voxel grids inside the eye window are input to the 3D CNN. If the side lengths of the eye window are $N_1$, $N_2$ and $N_3$, respectively, there are $N1\times N2\times N3$ voxel grid units within it. We use the filters to convolve the units, and a 3D feature matrix is obtained after every convolution operation is performed. Every convolutional layer has three parameters, \ie the number, stride and size of the filter. The convolution in the $n$-th layer can be expressed as $Conv_n(f; d; s)$, which indicates $f$ filters of size $d$ and at stride $s$. The exact parameters of the three layers in our model is $Conv_1(8; 5; 3); Conv_2(16; 4; 2); Conv_3(32; 3; 1)$. After the three convolutions, multiple feature matrices are obtained. Through average pooling, we get a fully-connected layer, and the layer is input to two more fully-connected layers. After an activation function $softmax$, a $1*2$ vector $[R_1,R_2]$ is obtained. We call it a reward vector since it is used to calculate the reward value (or the probability of the class object within the eye window) in the following procedures. When the points of the class object are located in the eye window, the value of the first dimension increases and the second decreases, \eg $[1,0]$, and vice versa. It indicates the probability of the eye window containing the points which belong to the class object.

\subsubsection{Object Class Detection by the DQN}

The 3D CNN predicts the probability of the points in the eye window belonging to a certain class rather than labels every point. It would only produce a high probability when the exact position and boundary of the class object are determined. The process for searching the objects can be expressed as the following optimization problem:
\begin{equation}
  E(p_b,q_b)^*=\argmax(f(E(p,q)))
\end{equation}
Our goal is to estimate a position $p_b$ and size $q_b$ of the eye window in the scene to
maximize $f(E(p,q))$.

During the searching process, the eye window produces a situation under every state. If the situation does not meet the criterion, the eye window needs to readjust its position and size to get a new state. An interaction happens between the movement behavior of the eye window and the step-by-step process of the sequential decision making on a discrete time series.

Based on the analysis, we design a 3DCNN-DQN dynamic searching mechanism: the 3D CNN is employed to evaluate the current state and the result, \ie probability of eye window containing the points is then sent to the DQN to instruct the next action of the eye window. Through the $evaluation \& feedback \xrightarrow { } action \xrightarrow { } evaluation \& feedback \xrightarrow { } action$ pattern, the eye window can localize the points of a certain class effectively. Figure 3 shows an example of the searching process, and a video recording the searching process can be found in our supplementary materials.

To quantify this mechanism, in the following we define $Q$ to represent the value of a state. At the same time, we explain how to choose an action based on $Q$ and how to update the parameters of the DQN. Based on Prioritized Replay and Dueling Network brought up by DeepMind in 2015 and 2016 respectively, we improve the architecture of the DQN for enhancing the efficiency of the point cloud parsing.
  \begin{figure}
  \begin{center}
  \subfigure{
  \begin{minipage}{4cm}
   \includegraphics[width=1.04\linewidth]{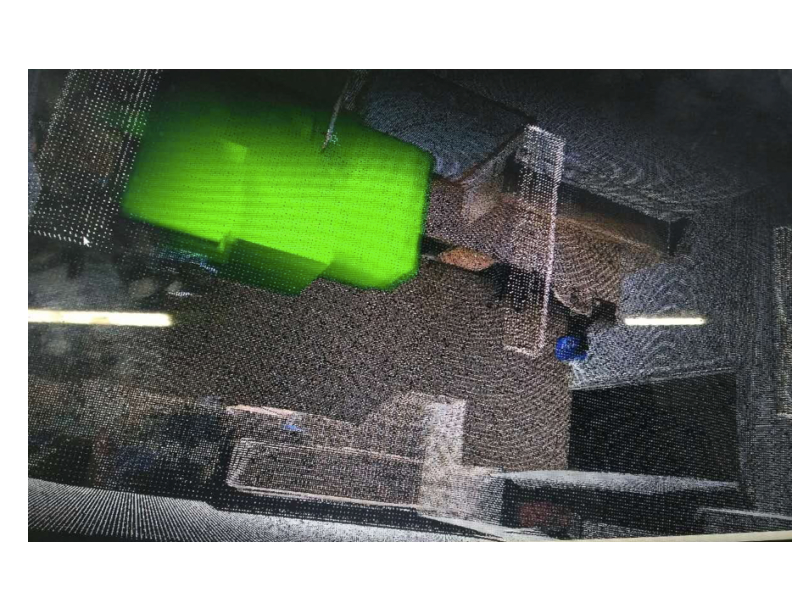}
   \end{minipage}}
   \subfigure{
  \begin{minipage}{4cm}
   \includegraphics[width=0.93\linewidth]{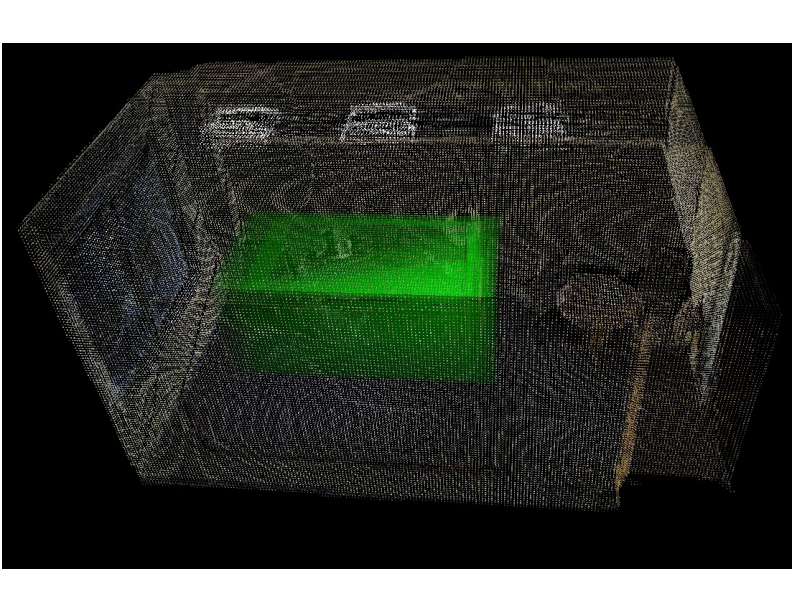}
   \end{minipage}}
  \end{center}
  \caption{Two angles of eye window searching tables. Locked as marked green. }
  \label{fig:long}
  \label{fig:onecol}
  \end{figure}

\paragraph{a. $Q$ - Define Value and Advantage} 
The 3D CNN is a hierarchical perceptive system whose parameters do not change after the training process. When the DQN is applied to localize the data in the eye window, the current state (\ie points within eye window) is input to the DQN. If the DQN uses the same convolutional layers as the 3D CNN, it cannot back propagate parameters to the convolutional layers. To make the parameters adjustable, we use the same structure of the convolutional layers as the 3D CNN, but the parameters are independent from the convolutional layers. The 3D CNN acts as the input channel of the current state for the DQN. Using convolutional networks to provide state of the environment for Reinforcement Learning Network has been proved to be very efficient~\cite{mnih2015human,mnih2013playing}. After three convolutions, the vector obtained by the fully connected layers is copied into two independent streams. The first stream outputs a scalar $V(s;\theta,\beta)$ through several fully-connected layers. The second stream outputs $\|A\|$-dimensional vector or $A(s, a; \theta,\alpha)$ through the fully-connected layers whose number is the same as the first stream, where $s$ represents the state of the eye window, $\theta$ represents the parameters of the convolutional layers, $\alpha$ and $\beta$ represent parameters of each stream, respectively. So, $Q$ is defined as the action $a$ value under the status $s$.
   \begin{equation}
   Q(s,a;\theta,\alpha,\beta)=V(s;\theta,\beta)+A(s,a';\theta,\alpha)-\frac{\sum_{a'}A(s,a;\theta,\alpha)}{\|A\|}
   \end{equation}
\paragraph{b. Policy - How to Act}
An agent chooses an action from a discrete set that can be presented as $a = [p_1,p_2,p_3,p_4,p_5,p_6]$ where $p_k\in \{-1,0,1\}, k=1,2,...,6$, which means the eye window has six sides and each side can choose three actions: expand one unit, contract one unit and static, thus it has $18$ variances at a time. The actions are given in the formula $a*=\argmax Q(s,a ; \theta_t)$. That is, the action which can obtain the largest $Q$ will be selected. The eye window can directly obtain a $Q$ under any state or action. In a traditional DQN, a considerable amount of time and space are needed to store the state-action pairs for later use~\cite{mnih2013playing}. In our method, we only store the size and position of a state, which makes the network have a higher computation efficiency.

The purpose of the eye window is to get the largest expected return. A multiple following-step policy is a better choice for getting the return. Therefore, the DQN will simulate $N$ following steps, that is, $N$ times actions are conducted based on the largest $Q$, and an accumulated $Q$ value $Q_{tg}$ would be our optimization goal, which is obtained by the following equation:
\begin{equation}
Q_{tg}=\tanh (\sum_{t=0}^{N-1} \lambda^t r_t+\lambda^NQ')
\end{equation}
where $[R_1, R_2]$ is the reward vector; $r=2R_1+1-R_2$. $\lambda$ is the decay coefficient satisfying $\lambda \in [0,1]$; $Q'$ is the final $Q$ after $N$ times simulation. The goal of $Q'$ is to approach the limitation situation, which should be presented as  $\sum_{t=0}^{\infty} {\lambda}^{t} r_t$. The network parameters are updated when the current $Q$ has a gradient descent to $Q_{tg}$:
\begin{equation}
\theta _{T+1}=\theta _T+\lambda (Q_{tg}-Q(s,a;\theta _T))\nabla _{\theta _T}Q(s,a;\theta)
\end{equation}
The DQN parameters are refreshed after a complete $N$-step-simulation, \ie the DQN has a preliminary probe into the possible approaches within $N$ steps, and save this probe result as new parameters in the network for later decisions. We perform the simulations $k$ times to get $k$ times probe simulations and parameter update. An action is determined based on the largest $Q$ when the last parameter is updated.

\renewcommand{\algorithmicrequire}{\textbf{Input:}}
\renewcommand{\algorithmicensure}{\textbf{Output:}}

    \begin{algorithm}
        \caption{3DCNN-DQN Algorithm}
        \begin{algorithmic}
            \Require \\ max iterate step $mis$, max reside step $mrs$, max simulate step $mss$, decay rate $\lambda$, mark threshold $mth$, state-reward dictionary $rd$, replay memory $rm$, $Q$ value network $Q_{net}$, reward network $R_{net}$
                \State $Is\gets 0$  , $Rs\gets 0$, $Ss\gets 0$ ,$cS\gets initial$  $state$
                \While{$Is<mis$}
                   \State action $(a, Q)\gets Q_{net}(cS, a; W)$
                   \State $cS\gets environment(cS, a)$ , $Rs\gets 0$
                   \While{$Rs<mrs$}
                        \State $Ss\gets 0$, simulate state set $S\gets []$
                        \State reward set $R\gets []$, $S[Ss]\gets cs$
                        \While{$Ss<mss$}
                             \State action $(a, Q)\gets Q_{net}(S[Ss], a; W)$ or $rm$
                             \State $S[Ss+1]\gets environment(S[Ss], a)$
                             \State $Ss\gets Ss+1$
                             \If{$S[Ss]$ not in $rd's$ $key$ $set$}
                                 \State $r'\gets R_{net}(S[Ss])$
                                 \State $rd[S[Ss]] \gets r'$
                             \EndIf
                             \State $R[Ss-1] \gets rd[S[Ss]]$
                        \EndWhile
                        \State $Q_{tg} \gets \tanh (\sum_{t=0}^{N-1} \lambda^t r_t+\lambda^NQ)$
                        \State Update $Q$'s W using $Q_{tg}$
                        \State $Rs \gets Rs+1$
                    \EndWhile
                    \State $Is \gets Is+1$
                \EndWhile
        \end{algorithmic}
    \end{algorithm}

\paragraph{d. Random Walk and Winner Replay}
Actions cause the movement and scaling of the eye window because of the above-mentioned steps. Meanwhile, $Q$ and the network parameters are also updated. However, pitfalls may happen when the eye window refuses to move as a failure in $Q$ value increases after $3^6$ dimensional actions. This situation happens occasionally in the local space. The eye window may get stuck in loops because of a combination of several spaces. We design a random walk mechanism to avoid this kind of pitfall. The eye window selects an action randomly in such a situation. As the eye window returns back to a position it ever reached, we present a mechanism named winner replay to recall its previous state. $Q$ and $Q_{tg}$ are simultaneously obtained when the eye window does $N$-steps simulations. $k$ numbers of $Q_{tg}$ are produced after $k$-times $N$-th step simulations. The smallest $Q_{tg}$ which is the result of $\|Q-Q_{tg}\|$ will be the winner of this particular state. Every time the eye window is under the same state, half of the probability is given for action selection based on the winner action, and half of the probability is given for action selection based on the normal $N$-step simulation. Experiments show that efficiency of the DQN is much improved by the winner replay mechanism, and the searching time is reduced by $73\%$ on average.

\paragraph{e. Lock the Target}
As mentioned above, the 3D CNN can be regarded as a confidence function deciding whether the eye window envelops the class objects. If the threshold value of the reward is not less than $0.9$, we conclude that the eye window has localized the points belonging to the class object. The points in the eye window are labelled.

Then, we compute the feature of the points within the eye window from every layer of the 3D CNN. The feature matrix of every layer is fully connected and concatenated, \eg the output of the first convolutional layer is fully-connected to get a feature vector $f_1$. Similarly, we get $f_2$ and $f_3$ from the second layer and third layer. The concatenated vector $[f_1, f_2, f_3]$ is the encoded feature for all the points within the eye window and is called the 3D CNN feature. 

Due to the point cloud rasterization, lots of valuable information about the shape and geometric layout of objects is lost. To make up the deficiency of the 3D CNN, the 3D CNN feature combined with the points in each eye window, \ie $[x, y, z, r, g, b, f_1, f_2, f_3]$, is taken as the input of Network 2.


\subsection{Network 2: Residual RNN for Meticulous Parsing}
The RNN further learns the features of the points in the eye window. The sequence of the points which is input to the RNN can be taken as a hidden Markov chain. When the points are input according to its spatial arrangement, the RNN can recognize the connection and difference between the features of multiple scales. These features are fused or abstracted. However, point cloud data is unordered and the spatial information it carries can be extremely complicated to unscramble. In order to fully simulate the hidden Markov chain, the RNN should be deep enough and contain enough number of parameters to fit the corresponding nonlinear transition function. We build a multiple layer Residual RNN to meet the requirement. LSTM cell is used to prevent gradient vanishing and enables the network to have a long-term memory. Residual Block is used to prevent degradation of the deep network.

\subsubsection{Residual RNN Structure}
Every point $P_k$ in the eye window corresponds to a reconstructed vector $V_k = [x_k, y_k, z_k, r_k, g_k, b_k, f_1, f_2, f_3]$, where $x_k, y_k, z_k$ are the coordinate of $P_k$; $r_k, g_k, b_k$ are the color of $P_k$; $f_1, f_2, f_3$ are the 3D CNN feature vector of every point in the eye window. After all reconstructed vectors within the eye window are obtained, we input them into the Residual RNN following the original spatial arrangement for training. The used Residual RNN has $7$ fully-connected layers, $3$ dropout layers, $2$ residual blocks and a LSTM cell as shown in Figure 4.

   \begin{figure}
   \begin{center}
      \includegraphics[width=0.8\linewidth]{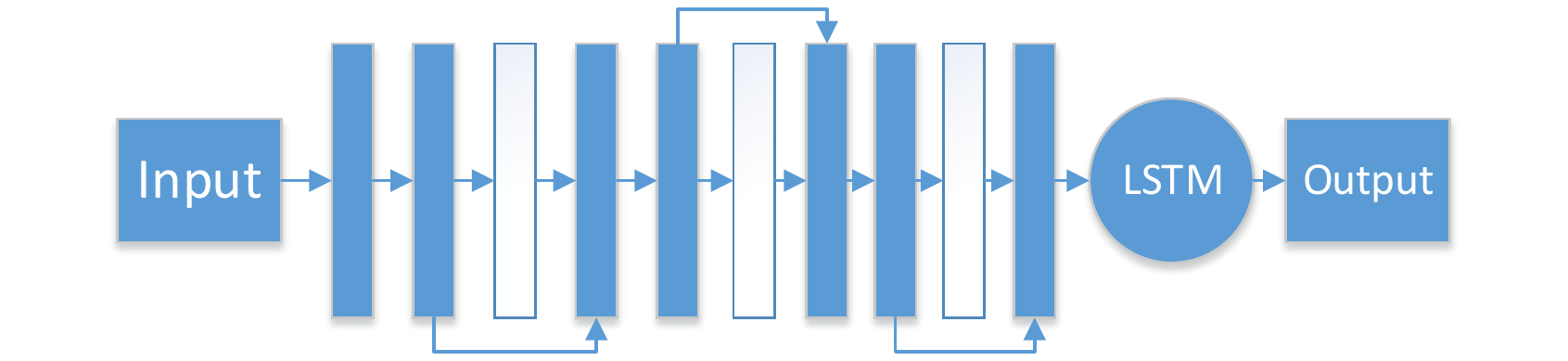}
   \end{center}
      \caption{The inner structure of Network 2. Blue rectangle means fully-connected layers, white means drop layer.}
   \end{figure}

\subsubsection{LSTM Cell}
The points in each eye window are still large. Thus the point sequence that is input to the Residual RNN is long. Network 2 is required to have the ability to understand the context among points on a large scale. The LSTM cell enables the network to learn long-term dependency of a sequence~\cite{hochreiter1997long}, \ie the connections and differences of the features of points.

\subsubsection{Residual Block}
A too deep network has a degradation problem, and performance of the network decreases as the layers become deeper. In the point cloud parsing, a deep network is necessary, but its too deep depth can easily cause a high training error which is called degradation~\cite{he2016deep}. Inspired by the Deep Residual Network~\cite{he2016deep}, we utilize the structures of the residual block to make some overlap joints among cells for solving the degradation.

\subsubsection{Deep RNN - A Multi-Layer Classifier}
Due to the fact that the feature vectors obtained by Network 1 derive from different types and scales of objects, the features should be efficiently fused in a high dimensional space. If the feature vectors are fed into a classifier like SVM or Random Forest, the self-adjustment of the feature representations may be restricted. That is to say, these classifiers are relatively shallow models and may not suit our problem. Our proposed multi-layer neural network can learn discriminative feature representations about locations, spatial relationship and color of the points, and fuse the features well to achieve high-quality classification results. 

\section{Experiments}

\begin{table*}
\small
\renewcommand{\arraystretch}{1} 
\begin{tabular}{ l | c  c  c  c  c  c  c  c  c  c  c  c | r }
  \hline  & ceiling & floor & wall & beam & column & window & door & table & chair & sofa & bookcase & board & mean\\
  \hline N1+N2 & 89.64 & 95.02 & 60.08 &  \textbf{78.55} & 89.36 & \textbf{75.29} & 33.41 & 70.48 & 58.14 & 76.98 & \textbf{84.97} & 37.21 & \textbf{70.76}\\
  \hline S1 & 71.61 & 88.70 & \textbf{72.86} & 66.67 & \textbf{91.77} & 25.92 & 54.11 & 46.02 & 16.15 & 6.78 & 54.71 & 3.91 & 49.93\\
  \hline N1 & \textbf{96.97} & \textbf{100.00} & 24.10 & 12.74 & 27.40 & 32.88 & \textbf{91.48} & \textbf{77.41} & 50.53 & \textbf{87.17} & 53.84 & 32.83 & 57.28 \\
  \hline N2 & 87.51 & 97.66 & 27.45 & 32.96 &  3.90 & 67.27 & 15.16 & 10.77 & \textbf{68.17} & 68.47 & 12.91 & \textbf{43.56} & 36.63 \\
      \hline FCN-12 & 20.32 & 11.61 & 8.34 & 12.69 &  33.36 & 20.41 & 10.01 & 9.68 & 11.27 & 2.24 & 1.89 & 13.86 & 12.97 \\
    \hline FCN-6 & - & 46.58 & 12.97 & - &  - & 29.55 & 11.65 & 50.96 & - & - & 20.32 & - & 28.64 \\
    \hline FCN-1 & - & 87.62 & - & 55.34 &  - & 63.46 & 30.68 & - & 60.27 & - & 80.37 & - & 62.96 \\
    \hline S2 & - & - & - & - & -  & - & - & 46.67 &33.80 & 4.76 & - & 11.72 &  24.24\\
\hline
\end{tabular}
\caption{N1+N2 is our method. S1, S2 correspondingly refer to the method of Armeni \emph{et al.}~\cite{armeni20163d} and Qi \emph{et al.}~\cite{qi2016pointnet}}
\label{tab:1}
\end{table*}

We train the 3D CNN and RNN on two NVIDIA K40 GPUs. The proposed method is implemented with Python and TensorFlow which can fully consume the calculating power in two GPUs. 

The first used dataset is the Stanford 3D semantic parsing data set~\cite{armeni20163d}. This dataset contains 3D scans from Matterport scanners in 6 areas including 271 rooms with a total of 6020 square meters. It has been fully annotated for 12 semantic classes which are structural elements (ceiling, floor, wall, beam, column, window and door) and commonly found items and furniture (table, chair, sofa, bookcase and board). We randomly choose 70\% rooms in every area as the training set, and the rest are taken as the testing set. When the proposed method is applied on the training set, we find that some fixed and relatively plain structures like ceiling, wall and floor can be easily parsed without using Network 1, so we feed them directly into Network 2. After a supervised learning process of the Residual RNN (Network 2), we feed all points of the scene into it. The  Residual RNN parses the points of three classes: ceiling, wall, floor from the whole point cloud. We then apply our method on the rest points to label the points of other classes. A part of parsing result is shown in Figure 5.
\begin{figure}
  \begin{tabular}{cc}

\subfigure[ground truth]{
\begin{minipage}{2.6cm}
  \includegraphics[width=0.9\linewidth]{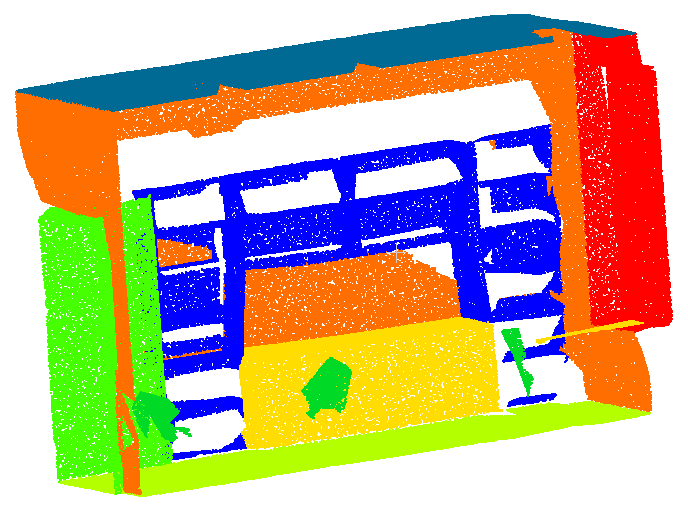}
  \end{minipage}}

  \subfigure[prediction]{
\begin{minipage}{2.6cm}
  \includegraphics[width=0.9\linewidth]{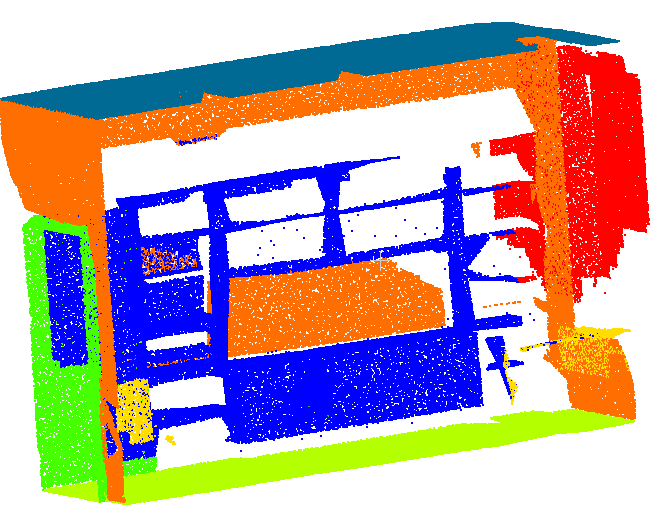}
 \end{minipage}
 }

 \subfigure[result]{
\begin{minipage}{2.6cm}
 \includegraphics[width=0.9\linewidth]{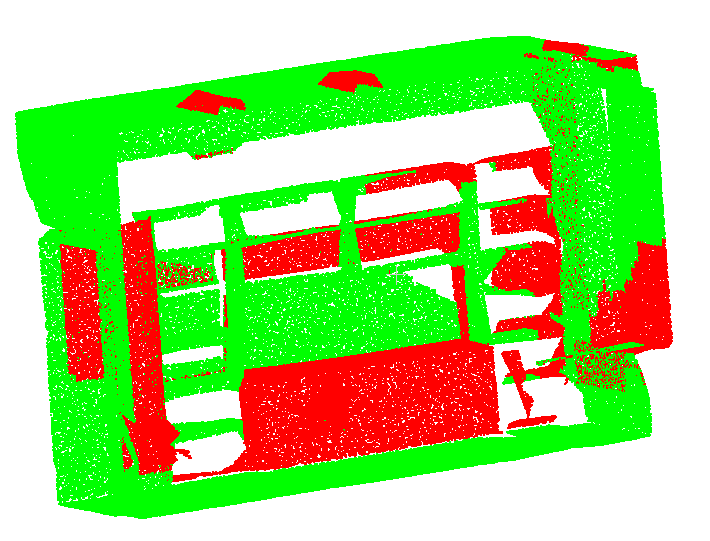}
\end{minipage}
}
\end{tabular}
  \caption{ Parsing result of a room. Green part in (c) means predicted right and red means wrong.}
\end{figure}

To validate the performance of our method in terms of point cloud parsing accuracy, we individually use Network 1 and Network 2 to classify the above dataset. We also compare the classification results with those obtained by using the methods of Armeni \emph{et al.}\cite{armeni20163d} and Qi \emph{et al.}~\cite{qi2016pointnet}. The classification results are listed in Table 1. 

The second used 3D point cloud data comes from the SUNCG dataset~\cite{song2016ssc}.  The SUNCG dataset contains 45,622 house models, 2,549 object files with corresponding material files and many texture files. It  has been annotated for 84 classes. We convert the data into 3D cloud points. Each point in the point cloud of each house model has an attribute $(X; Y; Z; R; G; B)$. $X, Y$ and $Z$ are the coordinate of the point, and $R, G$ and $B$ are the color of the point. The classification results can bee seen in our supplementary materials.

\begin{figure*}
\begin{center}
  \includegraphics[width=1.0\linewidth]{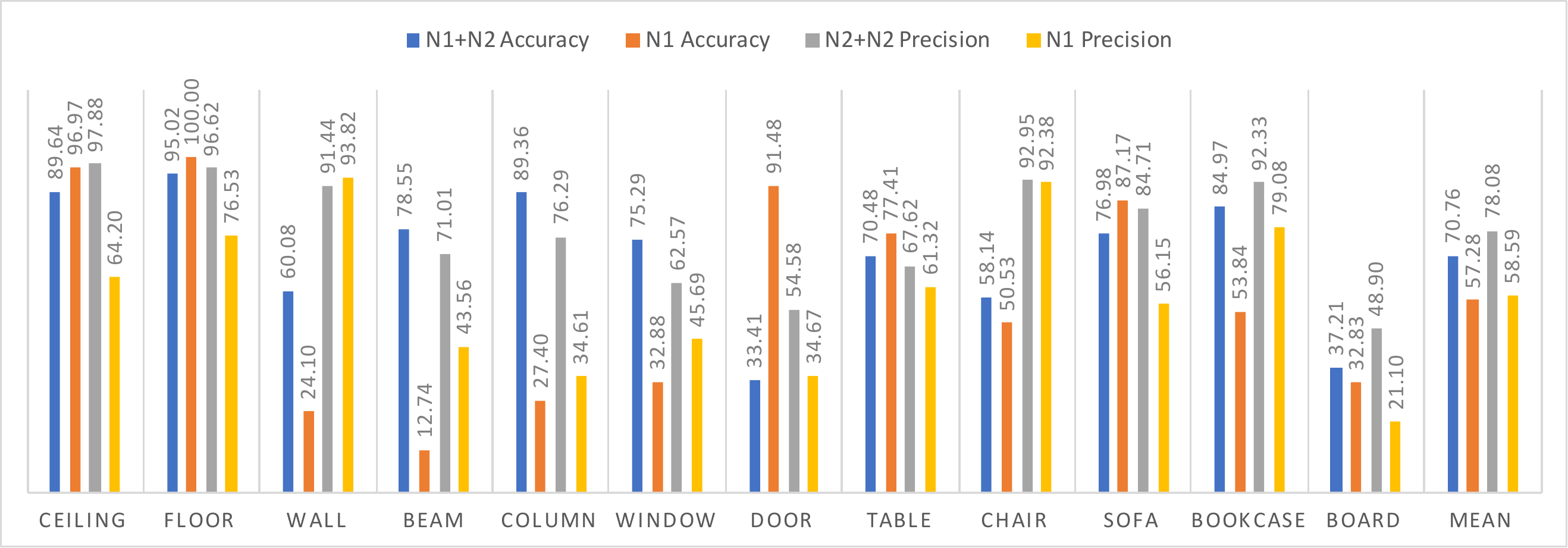}
\end{center}
\caption{Accuracy and Precesion of our method. N1, N2 represent Network 1 and Network 2.}
\label{fig:long}
\end{figure*}

\subsection{Comparison of Our Method with the FCN}
The FCN used in image processing may seem to function similarly with our framework since the FCN also does an end-to-end translation that gives each pixel an individual label. However, for large-scale 3D scenes, the spatial information can be extremely complex. We expand the classic VGG-16 model~\cite{simonyan2014very} into a 3D FCN, and compare it with our method based on the same training set and testing set. 

The classification results are listed in Table 1. From this table, it is noted that the 3D version of the FCN achieves poor performance compared with our method. Point cloud data is quite different from image-based 2.5D data like RGBD images. Its large volume, noise, point density inhomogeneity and disorder make it hard to draw the connections among points. For achieving high-quality classification results of large-scale point clouds using the 3D FCN, a deeper network is required, which consumes much more computational cost and time. In our method, the eye window under the control of the DQN can lock the objects more accurately and efficiently. 

FCN-1, FCN-6 and FCN-12 mean to classify just one class, six classes and twelve classes at a time, respectively. FCN-1 fails to recognize the border, and the overall accuracy is not satisfactory. In FCN-6 and FCN-12, the accuracy reduces to a very low level. Furthermore, in FCN-12, the loss function doesn't converge.

\subsection{RNN Helps to Raise Precision}
As listed in Table 1, if only Network 1 is applied to parse the point cloud, the classification accuracy of the wall class is much lower than that obtained by using our method. The reason lies in that, in our experiment, the wall class is parsed later than most of other classes, which makes many of the wall points mis-classified by the eye window before they are segmented for other classes. For the classes like window, table, board, their classification accuracies obtained by Network 1 are little influenced. Accuracies of some classes are even higher, but there are relatively low precisions which decrease the classification accuracies of the classes that are parsed later. Usually, when the objects of two different classes are extremely close, like a table and a bookcase that are both closely leaned on a wall, and a chair that is under a table, the eye window is very likely to contain points of irrelevant classes. Network 2 helps to increase the precision of the two classes and also enhance classification accuracy of the classes that are parsed later through putting the wrong-included points back into the scene as shown in Figure 6.

\subsection{Influences of the DQN on Point Cloud Parsing}
We find that the classification accuracy heavily depends on how long we run the DQN. We set 5000 iterations for every class in a room. The DQN takes 13 hours to localize all 12 classes on average in the first dataset. If the localization process goes longer, the accuracy usually raises since the eye window has more time to observe more states of the room. According to the activity thermodynamic diagram (Figure 7) of an eye window, most of the time the eye window moves around the objects, but there are also areas which are not worth looking at. During the training of the DQN, it takes a long time for the eye window to eventually localize the class objects. In future work, we plan to design a new reinforcement learning strategy to make the localization process more efficient.

\subsection{3D CNN is Not Well Trained}

The main limitation of our method is that the 3D CNN is not trained well enough. For the applications of the DQN like playing games~\cite{mnih2015human,mnih2013playing}, the feedback of an action is structurally fixed and always accurate since it's directly given by the environment. For object detection from images, CNNs can be very efficient since it is trained on large-scale datasets~\cite{girshick2015fast}. The 3D CNN is hard to learn discriminative features from small point clouds. In this situation, the eye window often gets confused when it encounters a small bookcase that resembles a table or it often mis-classifies a part of a door made of glass into a window. If the 3D CNN cannot give an accurate enough reward, the eye window may be rude as it locks an object. In the experiment, the boundary of the eye window is not always accurate. For instance, when it locks a window it would mistakenly take part of the wall around it. Based on the facts, we believe that training the 3D CNN on a large dataset can help to enhance the classification performance.
\begin{figure}[H]
\begin{tabular}{cc}

\subfigure[table]{
\begin{minipage}{4cm}
   \includegraphics[width=0.9\linewidth]{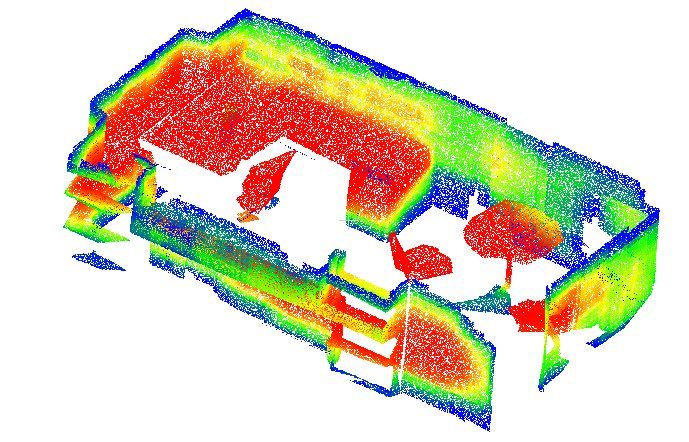}
   \end{minipage}}

   \subfigure[bookcase]{
\begin{minipage}{4cm}
   \includegraphics[width=0.9\linewidth]{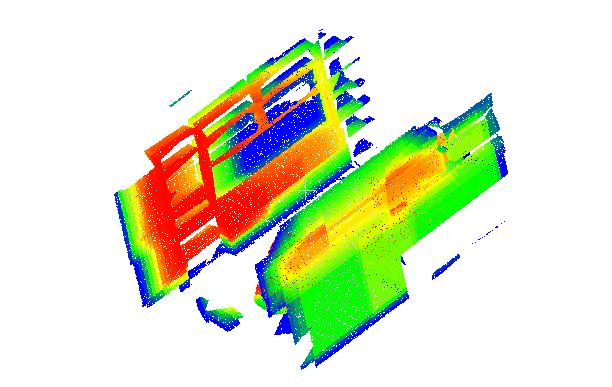}
   \end{minipage}}
\end{tabular}
   \caption{ Activity thermodynamic diagram of an eye window.}
\end{figure}

\section{Conclusion}
In this paper, we propose the 3DCNN-DQN-RNN framework for automatically parsing large-scale 3D point clouds. The 3D CNN has the ability to learn the features about spatial distribution shapes, colors and contexts of the points in each voxel grid unit from multi-scales, and fuse the features into a 3D CNN feature representation. The eye window under the control of the DQN can efficiently localize the class objects according to the reward. For further achieving high-quality classification results, the 3D CNN feature and the point coordinate and color of each point in the eye window are concatenate into one vector which is taken as the input of the Residual RNN. The output of the Residual RNN is the parsing result of the points in each eye window. Our framework also supports object localization, detection and retrieval.

 In future work, we will introduce the asynchronous advantage actor critic (A3C) to coordinate multiple eye windows for deriving the class objects more efficiently.

\section{Acknowledgments}
The work was supported by the National Natural Science Foundation of China under Grant 41371324.

{\small
\bibliographystyle{ieee}
\bibliography{egbib}

\begin{thebibliography}{10}\itemsep=-1pt

\bibitem{armeni20163d}
I.~Armeni, O.~Sener, A.~R. Zamir, H.~Jiang, I.~Brilakis, M.~Fischer, and
  S.~Savarese.
\newblock 3d semantic parsing of large-scale indoor spaces.
\newblock {\em In Proceedings of the IEEE Conference on Computer Vision and
  Pattern Recognition,}, 2016.

\bibitem{chehata2009airborne}
N.~Chehata, L.~Guo, and C.~Mallet.
\newblock Airborne lidar feature selection for urban classification using
  random forests.
\newblock {\em International Archives of Photogrammetry, Remote Sensing and
  Spatial Information Sciences}, 38(Part 3):W8, 2009.

\bibitem{chen2014semantic}
L.-C. Chen, G.~Papandreou, I.~Kokkinos, K.~Murphy, and A.~L. Yuille.
\newblock Semantic image segmentation with deep convolutional nets and fully
  connected crfs.
\newblock {\em International Conference on Learning Representations (ICLR)},
  2015.

\bibitem{fukushima1980neocognitron}
K.~Fukushima.
\newblock Neocognitron: A self-organizing neural network model for a mechanism
  of pattern recognition unaffected by shift in position.
\newblock {\em Biological cybernetics}, 36(4):193--202, 1980.

\bibitem{girshick2015fast}
R.~Girshick.
\newblock Fast r-cnn.
\newblock {\em Proceedings of the IEEE International Conference on Computer
  Vision}, pages 1440--1448, 2015.

\bibitem{guan2015deep}
H.~Guan, Y.~Yu, Z.~Ji, J.~Li, and Q.~Zhang.
\newblock Deep learning-based tree classification using mobile lidar data.
\newblock {\em Remote Sensing Letters}, 6(11):864--873, 2015.

\bibitem{he2016deep}
K.~He, X.~Zhang, S.~Ren, and J.~Sun.
\newblock Deep residual learning for image recognition.
\newblock pages 770--778, 2016.

\bibitem{hochreiter1997long}
S.~Hochreiter and J.~Schmidhuber.
\newblock Long short-term memory.
\newblock {\em Neural computation}, 9(8):1735--1780, 1997.

\bibitem{johnson1999using}
A.~E. Johnson and M.~Hebert.
\newblock Using spin images for efficient object recognition in cluttered 3d
  scenes.
\newblock {\em IEEE Transactions on pattern analysis and machine intelligence},
  21(5):433--449, 1999.

\bibitem{koppula2011semantic}
H.~S. Koppula, A.~Anand, T.~Joachims, and A.~Saxena.
\newblock Semantic labeling of 3d point clouds for indoor scenes.
\newblock {\em Advances in Neural Information Processing Systems}, pages
  244--252, 2011.

\bibitem{kragh2015object}
M.~Kragh, R.~N. J{\o}rgensen, and H.~Pedersen.
\newblock Object detection and terrain classification in agricultural fields
  using 3d lidar data.
\newblock {\em International Conference on Computer Vision Systems}, pages
  188--197, 2015.

\bibitem{lafarge2012creating}
F.~Lafarge and C.~Mallet.
\newblock Creating large-scale city models from 3d-point clouds: a robust
  approach with hybrid representation.
\newblock {\em International journal of computer vision}, 99(1):69--85, 2012.

\bibitem{lecun2015deep}
Y.~LeCun, Y.~Bengio, and G.~Hinton.
\newblock Deep learning.
\newblock {\em Nature}, 521(7553):436--444, 2015.

\bibitem{long2015fully}
J.~Long, E.~Shelhamer, and T.~Darrell.
\newblock Fully convolutional networks for semantic segmentation.
\newblock pages 3431--3440, 2015.

\bibitem{martinovic20153d}
A.~Martinovic, J.~Knopp, H.~Riemenschneider, and L.~Van~Gool.
\newblock 3d all the way: Semantic segmentation of urban scenes from start to
  end in 3d.
\newblock {\em Proceedings of the IEEE Conference on Computer Vision and
  Pattern Recognition}, pages 4456--4465, 2015.

\bibitem{maturana20153d}
D.~Maturana and S.~Scherer.
\newblock 3d convolutional neural networks for landing zone detection from
  lidar.
\newblock {\em IEEE International Conference on Robotics and Automation
  (ICRA)}, pages 3471--3478, 2015.

\bibitem{maturana2015voxnet}
D.~Maturana and S.~Scherer.
\newblock Voxnet: A 3d convolutional neural network for real-time object
  recognition.
\newblock {\em Intelligent Robots and Systems (IROS), 2015 IEEE/RSJ
  International Conference on}, pages 922--928, 2015.

\bibitem{mnih2013playing}
V.~Mnih, K.~Kavukcuoglu, D.~Silver, A.~Graves, I.~Antonoglou, D.~Wierstra, and
  M.~Riedmiller.
\newblock Playing atari with deep reinforcement learning.
\newblock {\em NIPS Deep Learning Workshop}, 2013.

\bibitem{mnih2015human}
V.~Mnih, K.~Kavukcuoglu, D.~Silver, A.~A. Rusu, J.~Veness, M.~G. Bellemare,
  A.~Graves, M.~Riedmiller, A.~K. Fidjeland, G.~Ostrovski, et~al.
\newblock Human-level control through deep reinforcement learning.
\newblock {\em Nature}, 518(7540):529--533, 2015.

\bibitem{qi2016pointnet}
C.~R. Qi, H.~Su, K.~Mo, and L.~J. Guibas.
\newblock Pointnet: Deep learning on point sets for 3d classification and
  segmentation.
\newblock {\em Proceedings of 30th IEEE Conference on Computer Vision and
  Pattern Recognition}, 2017.

\bibitem{qi2016volumetric}
C.~R. Qi, H.~Su, M.~Nie{\ss}ner, A.~Dai, M.~Yan, and L.~J. Guibas.
\newblock Volumetric and multi-view cnns for object classification on 3d data.
\newblock pages 5648--5656, 2016.

\bibitem{serre2005object}
T.~Serre, L.~Wolf, and T.~Poggio.
\newblock Object recognition with features inspired by visual cortex.
\newblock {\em 2005 IEEE Computer Society Conference on Computer Vision and
  Pattern Recognition (CVPR'05)}, 2:994--1000, 2005.

\bibitem{simonyan2014very}
K.~Simonyan and A.~Zisserman.
\newblock Very deep convolutional networks for large-scale image recognition.
\newblock {\em International Conference on Learning Representations (ICLR)},
  2015.

\bibitem{song2014sliding}
S.~Song and J.~Xiao.
\newblock Sliding shapes for 3d object detection in depth images.
\newblock pages 634--651. Springer, 2014.

\bibitem{song2016deep}
S.~Song and J.~Xiao.
\newblock Deep sliding shapes for amodal 3d object detection in rgb-d images.
\newblock pages 808--816, 2016.

\bibitem{song2016ssc}
S.~Song, F.~Yu, A.~Zeng, A.~X. Chang, M.~Savva, and T.~Funkhouser.
\newblock Semantic scene completion from a single depth image.
\newblock {\em Proceedings of 30th IEEE Conference on Computer Vision and
  Pattern Recognition}, 2017.

\bibitem{wang2015multiscale}
Z.~Wang, L.~Zhang, T.~Fang, P.~T. Mathiopoulos, X.~Tong, H.~Qu, Z.~Xiao, F.~Li,
  and D.~Chen.
\newblock A multiscale and hierarchical feature extraction method for
  terrestrial laser scanning point cloud classification.
\newblock {\em IEEE Transactions on Geoscience and Remote Sensing},
  53(5):2409--2425, 2015.

\bibitem{wu20153d}
Z.~Wu, S.~Song, A.~Khosla, F.~Yu, L.~Zhang, X.~Tang, and J.~Xiao.
\newblock 3d shapenets: A deep representation for volumetric shapes.
\newblock {\em Proceedings of the IEEE Conference on Computer Vision and
  Pattern Recognition}, pages 1912--1920, 2015.

\bibitem{yang2017computing}
B.~Yang, Z.~Dong, Y.~Liu, F.~Liang, and Y.~Wang.
\newblock Computing multiple aggregation levels and contextual features for
  road facilities recognition using mobile laser scanning data.
\newblock {\em ISPRS Journal of Photogrammetry and Remote Sensing},
  126:180--194, 2017.

\bibitem{yu2015multi}
F.~Yu and V.~Koltun.
\newblock Multi-scale context aggregation by dilated convolutions.
\newblock {\em International Conference on Learning Representations (ICLR)},
  2016.

\bibitem{zhang2006pixel}
L.~Zhang, X.~Huang, B.~Huang, and P.~Li.
\newblock A pixel shape index coupled with spectral information for
  classification of high spatial resolution remotely sensed imagery.
\newblock {\em IEEE Transactions on Geoscience and Remote Sensing},
  44(10):2950--2961, 2006.

\bibitem{zhang2013tensor}
L.~Zhang, L.~Zhang, D.~Tao, and X.~Huang.
\newblock Tensor discriminative locality alignment for hyperspectral image
  spectral--spatial feature extraction.
\newblock {\em IEEE Transactions on Geoscience and Remote Sensing},
  51(1):242--256, 2013.

\bibitem{zhang2016multilevel}
Z.~Zhang, L.~Zhang, X.~Tong, P.~T. Mathiopoulos, B.~Guo, X.~Huang, Z.~Wang, and
  Y.~Wang.
\newblock A multilevel point-cluster-based discriminative feature for als point
  cloud classification.
\newblock {\em IEEE Transactions on Geoscience and Remote Sensing},
  54(6):3309--3321, 2016.

\bibitem{zhou2013complete}
Q.-Y. Zhou and U.~Neumann.
\newblock Complete residential urban area reconstruction from dense aerial
  lidar point clouds.
\newblock {\em Graphical Models}, 75(3):118--125, 2013.

\end{thebibliography}
}

\clearpage



\section*{Supplementary material}

# 3DCNN-DQN-RNN: A Deep Reinforcement Learning Framework for Semantic Parsing of Large-scale 3D Point Clouds: Supplementary Materials

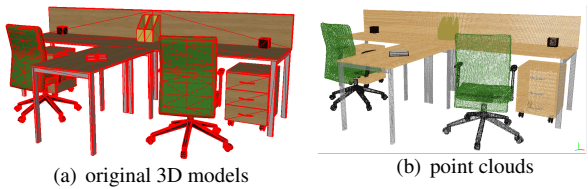

(a) original 3D models  (b) point clouds

*Figure 1: 3D models and their point clouds. (a) 3D models of tables and its neighboring chairs. (b) the corresponding point cloud data.*

## A. Overview

The supplementary material provides a demo video and an additional quantitative test example to the main paper.

In Sec B, we provide a demo video to illustrate the process that an eye window localizes and detects the category table in a room of the Stanford 3D semantic parsing data set [1] (Stanford dataset). In Sec C, we give the classification result of the SUNCG point cloud data obtained by our method, and analyze the experimental result. The training dataset is composed of the Stanford dataset and 20% of the SUNCG point cloud data. The other SUNCG point cloud data is taken as the testing data. The SUNCG point cloud data, which is derived from the SUNCG dataset [2], has been described in Section 4 of the main paper. Fig. 1(a) illustrates 3D models of tables and its neighboring chairs. Fig. 1(b) illustrates their corresponding point cloud data.

To validate the performance of our method, we also obtain the classification result using the method of Armeni et al. [1] on the same training data and testing data as ours.

## B. A Demo Video of the Detection Process

The attached video shows the process of an eye window detect the category table in a room of the Stanford dataset. There are mainly two types of tables in the scene. As illustrated in the video, it is noted that the eye window can envelop all of the tables in the scene. We also observe that the DQN has the ability to rapidly localize the approximate locations of the tables, and spend most of time on detecting the boundaries of the tables. In the experiment, we only classify a point into a category if the point is detected by the eye window for more than 5 times. Even the boundaries seem rough in the video, the final classification performance is still high. The eye window has also spent a lot of time around the bookcase since the bookcase resembles a table in its height and structure. We plan to integrate the context information into our method to enhance the detection performance of the eye window.

## C. Experimental Results on SUNCG Point Cloud Dataset

### C.1. Training data and testing data

To further validate the performance of our method, the training data comes from two different indoor point clouds. We then transfer the trained parameters of our network model to the testing data for performing the classification. Specifically, we first train the 3D CNN on the Stanford dataset. Then, in order to make the 3D CNN adapt to the new environment, based on the original parameters we use 20% of the SUNCG point cloud data to further train the 3D CNN. The training process is the same as it was on the Stanford dataset. It costs 48 hours and the training error converges to approximately 10%. The other 80% of the SUNCG point cloud data are taken as the testing data.

### C.2. Comparison between our method and the method of Armeni *et al.* [1]

Table 1 lists the classification results obtained by our method and the method of Armeni *et al.* [1]. From this table, it is noted that our method outperforms the method of Armeni *et al.* [1].

|       | door  | table | bookcase | chair | mean  |
|-------|-------|-------|----------|-------|-------|
| N1+N2 | 18.53 | **52.31** | **27.01** | **34.98** | **33.21** |
| S1    | **21.84** | 29.60 | 14.14 | 8.66 | 18.56 |

*Table 1: Comparison of methods. N1+N2 is our method. S1 refers to the method of Armeni* et al. *[1].*

## C.3. Analysis of the experimental result

The classification performance of our method on the SUNCG point cloud dataset is not higher than that on the Stanford dataset. The main reasons lie in twofold.

On the one hand, the complexity and irregularity of objects in the SUNCG point cloud dataset. For instance, chairs in the Stanford 3D data set generally have two styles, and thus this similarity of structure among chairs can be easily learned by the 3D CNN. In this way, the 3D CNN produces more accurate scores for the DQN to determine the next action of the eye window. However, in the SUNCG point cloud dataset, the styles of chairs are different in different rooms. We lack enough training data with similar features to train the 3D CNN for producing an accurate score. The same cases happen to other objects. Compared with the Stanford dataset, the spatial relationship among objects in the SUNCG point cloud dataset are more complicated, and the patterns of the objects are also more difficult to be recognized.

On the other hand, the performance of the used computer hardware is limited. In order to enable the proposed method to deal with large-scale 3D point clouds, we utilize the 3D CNN with 6 layers (including convolutional and fully connected layers) to parse the point clouds. The too limited layers of the 3D CNN reduce the generalization ability of the 3D CNN. Due to the limitations of the hardware, fierce down-sampling is applied in our network model. In this situation, the feature representation is not discriminative. If a deeper 3D CNN with less down-sampling is employed to handle point clouds of complex scenes like the SUNCG point cloud dataset, we believe the classification accuracy would be greatly enhanced.

## References

[1] I. Armeni, O. Sener, A. R. Zamir, H. Jiang, I. Brilakis, M. Fischer, and S. Savarese. 3d semantic parsing of large-scale indoor spaces. *In Proceedings of the IEEE Conference on Computer Vision and Pattern Recognition,*, 2016. 1

[2] S. Song, F. Yu, A. Zeng, A. X. Chang, M. Savva, and T. Funkhouser. Semantic scene completion from a single depth image. *Proceedings of 30th IEEE Conference on Computer Vision and Pattern Recognition*, 2017. 1



\end{document}